%% file: main.tex
\documentclass{article}

\usepackage{microtype}
\usepackage{graphicx}
\usepackage{multirow}
\usepackage{soul}
\usepackage{booktabs} 
\usepackage{enumitem}
\usepackage{caption}
\usepackage{subcaption}
\usepackage{hyperref}

\usepackage[accepted]{icml2023}

\usepackage{amsmath}
\usepackage{amssymb}
\usepackage{mathtools}
\usepackage{amsthm}
\usepackage{bm, bbm}

\usepackage[capitalize,noabbrev]{cleveref}

\theoremstyle{plain}

\theoremstyle{definition}

\theoremstyle{remark}

\usepackage[textsize=tiny]{todonotes}

\renewcommand{\todo}[1]{{\color{red} todo: {#1}}}

\newcommand{\todobox}[3]{%
	\colorbox{#1}{\textcolor{white}{\sffamily\bfseries\scriptsize #2}}%
	~\textcolor{red}{#3} %
	\textcolor{#1}{$\triangleleft$}%
}
\newcommand{\notebox}[3]{%
	\colorbox{#1}{\textcolor{white}{\sffamily\bfseries\scriptsize #2}}%
	~\textcolor{blue}{#3} %
	\textcolor{#1}{$\triangleleft$}%
}
\newcommand{\addcite}[2]{%
	\colorbox{red}{\textcolor{white}{\sffamily\bfseries\scriptsize cite}}%
}

\renewcommand{\todo}[1]{\todobox{red}{todo}{#1}}

\newcommand{\DP}[1]{\notebox{pink}{DP}{#1}}

\newcommand{\MR}[1]{\notebox{teal}{MR}{#1}}

\newcommand{\SCClip}{\textsc{SCClip}}
\newcommand{\Clip}{\textsc{Clip}}
\newcommand{\adv}{\mathtt{b}}
\newcommand{\victim}{\mathtt{v}}
\newcommand{\honestUsers}{\mathcal{V}_\mathcal{H}}
\newcommand{\byzantineUsers}{\mathcal{V}_\mathcal{B}}

\icmltitlerunning{Can Decentralized Learning Be More Robust Than Federated Learning?}

\begin{document}

\twocolumn[
\icmltitle{Can Decentralized Learning be more robust than Federated Learning?}



\icmlsetsymbol{equal}{*}

\begin{icmlauthorlist}
\icmlauthor{Mathilde Raynal}{epfl}
\icmlauthor{Dario Pasquini}{epfl}
\icmlauthor{Carmela Troncoso}{epfl}
\end{icmlauthorlist}

\icmlaffiliation{epfl}{Spring Lab, EPFL, Switzerland}

\icmlcorrespondingauthor{Mathilde Raynal}{mathilde.raynal@epfl.ch}

\icmlkeywords{Machine Learning, ICML}

\vskip 0.3in
]



\printAffiliationsAndNotice{}  

\input{parts/0-abstract}
\input{parts/1-intro.tex}
\input{parts/2-background.tex}
\input{parts/3-attacks.tex}
\input{parts/4-experiments.tex}
\input{parts/5-DLvsFL.tex}
\input{parts/6-discussion.tex}

\bibliography{references}
\bibliographystyle{icml2023}

\newpage
\appendix
\onecolumn
\input{app/appendix.tex}

\end{document}

%% file: parts/0-abstract.tex
\begin{abstract}
Decentralized Learning (DL) is a peer--to--peer learning approach that allows a group of users to jointly train a machine learning model.
To ensure correctness, DL should be robust, i.e., Byzantine users must not be able to tamper with the result of the collaboration. 
In this paper, we introduce two \textit{new} attacks against DL where a Byzantine user can: 
make the network converge to an arbitrary model of their choice,
and exclude an arbitrary user from the learning process.
We demonstrate our attacks' efficiency against Self--Centered Clipping, the state--of--the--art robust DL protocol.
Finally, we show that the capabilities decentralization grants to Byzantine users result in decentralized learning \emph{always} providing less robustness than federated learning. 
\end{abstract}
\vspace{-0.7cm}

%% file: parts/1-intro.tex


\section{Introduction}
\label{sec:intro}

Collaborative Machine Learning (CML) allows users with private local training data to, as the name suggests, collaborate in training a machine learning model. 
Decentralized Learning (DL)~\cite{can} is a peer--to--peer approach to CML where users communicate with each other directly.
DL finds its use in applications where users wish to keep their local training data private.

In order to be useful, the output of the DL protocol---the model the users converge to---must be accurate and reliable, even in the presence of malicious participants.
No Byzantine user should be able to tamper with the result of the collaboration, e.g., coerce the users to converge to a final model of their choice, degrade the model performance, or force a poisoned model that benefits the Byzantine participant (e.g., containing a backdoor \cite{trojan}). 
Correctness in DL can typically be achieved through the use of robust aggregators~\cite{rfa}.



Existing attacks against DL prevent users from reaching consensus~\cite{scc} or poison the global model to cause misclassification of certain samples~\cite{labelflip}.
In this work, we show that the knowledge of other users' updates and the ability to send inconsistent updates, both enabled by decentralization, enables a Byzantine user in DL to perform harmful attacks that go beyond the adversarial objectives considered so far.
Concretely, we introduce two novel attacks that exploit the features of DL:
First, the \textit{state-override} attack, where the adversary is able to cancel honest users' contributions from the network and can then \textit{establish the model the network is going to converge to.}
Second, the \textit{sandtrap} attack where the adversary \textit{can isolate a user from the rest of the network} resulting on this victim not learning from the rest of the honest users, and the honest users not gaining anything from the participation of the victim in the learning process.

We demonstrate the effectiveness of our attacks against  \SCClip{}~\cite{scc}, the state-of-the-art robust DL algorithm. We show that despite \SCClip{}'s proof of robustness, it does not offer protection against Byzantine users.

Finally, we compare DL to Federated Learning (FL) through the robustness lens. We show that decentralization boosts the capabilities of Byzantine users with respect to the scenario in which interactions are mediated by a server. As a result, \textit{by design}, DL provides less robustness than FL, no matter how users are connected to each other, or whether they use robust aggregation.
We also show that this gap can only be reduced at the cost of efficiency and by constraining the decentralized topology.

To summarize, our contributions are as follow:
\vspace{-0.2cm}
\begin{itemize}[nosep]
    \item[\checkmark] We introduce novel Byzantine attacks against DL which achieve more harmful  objectives than attacks in the literature. 
    \item[\checkmark] We show that our attacks are effective against the state--of--the--art robust DL protocol~\cite{scc}.
    \item[\checkmark] We show that decentralization grants a number of advantages to Byzantine users with respect to learning proxied by a server. As a result, DL cannot provide better robustness than federated learning.
\end{itemize}
\vspace{-0.1cm}

%% file: parts/2-background.tex
\section{Background}
\label{sec:background}

\input{parts/2a-cl.tex}

\input{parts/2b-byz.tex}
\input{parts/2c-ra.tex}

%% file: parts/2a-cl.tex
Collaborative Machine Learning (CML) allows a set $\mathcal{V}$ of~$n$ users each with local data~$X_i$ to collectively optimize a model~$f$ with parameters $\theta \in \mathbb{R}^d$.
The objective of CML is to achieve \textit{average consensus} (also known as \textit{average agreement}, or simply \textit{consensus}), i.e., all users obtain and agree on a global model which is close to the average of honest users' models. 

\subsection{Decentralized Learning}

The decentralized approach to CML, Decentralized Learning (DL), proposes to achieve average consensus through gossip.
Each user $i$ communicates with a subset of users, referred to as its neighbors $\mathcal{N}_i$.
All communication links form a communication graph, that we refer to as the \textit{topology}.
Each link is associated with a weight. 
Those weights form a non--negative mixing matrix $\bm{W} \in \mathbb{R}^{n\times n}$.
Formally, $n$ users start with common model parameters~$\theta^0$ and iterate over the following three steps until a stop condition is met:
(1) \textit{Local training:} Users apply gradient descent on their local model parameters and compute a \textit{model update}~$\theta_{i}^{t+1/2}$.
(2) \textit{Communication:} Users share their model update $\theta_{i}^{t+1/2}$ with their neighbors $\mathcal{N}_i$, and receive their neighbors' updates. 
(3) \textit{Aggregation:} Users aggregate their neighbor's updates with their local one, and use the result of the aggregation to update their local state. 
Naive aggregation simply consists in computing the average of all received updates, i.e., $\theta^{t+1} = \frac{1}{|\mathcal{N}_i|}\sum_{i \in \mathcal{N}_i}\theta_i^{t+1/2}$.

The full DL training procedure is described in Algorithm \ref{alg:scc} in Appendix~\ref{sec:dl}. 

There are, in theory, no restrictions on how nodes connect to each other, but the DL literature assumes that the underlying graph is connected and that $\bm{W}$ is symmetric and doubly stochastic, i.e., $\forall i,j \in [n]: W_{ij} = W_{ji}, \sum_{i=1}^n\bm{W}_{ij}=1$, and $\sum_{j=1}^n\bm{W}_{ij}=1$.
The mixing matrix $\bm{W}$ is fixed before training and does not change.\looseness=-1

%% file: parts/2b-byz.tex
\subsection{Robustness attacks}
\label{sec:rw}

Since it is a possibility in CML that a participant is compromised, correctness of the output is essential.
These compromised participants can try to manipulate the output of the learning process and target the integrity property of CML.
Such attacks are referred to as \textit{robustness attacks}.
The literature of Byzantine attacks in CML is expansive, and the attacks can be categorized into two groups: attacks that aim to (A1) poison the output of the learning system in a specific way (e.g., misclassification of predefined samples)~\cite{history, backdoorfl, bhagoji, tails, dba, sun}; or (A2) cause general failure of the learning system (e.g., users converge to a model with poor utility)~\cite{alie, scc, ipm}.

%% file: parts/2c-ra.tex
\subsection{Robust aggregators}
Naive aggregation is not robust with regards to its input, i.e., it is permeable to irregularities (which can happen with training data of poor quality) and attacks (as the ones presented in Section~\ref{sec:rw}).
There exist more sophisticated aggregation functions, referred to as \textit{robust aggregators}, whose goal is to ensure a correct and accurate CML output, even in the presence of Byzantine users.
Robust aggregators limit (or fully discard) model updates that are believed to be detrimental to the learning process, e.g., updates that are far away from the mean. 
These robust aggregators are used by users during the aggregation step (3).\footnote{To be noted that, when using these robust aggregators, the convergence of the training process is not guaranteed, even if there is no Byzantine adversary~\cite{scc}.}

Most of the existing techniques can only deal with the scenario where less than half---or a third---of the participants are malicious~\cite{bridge, ubar, tm, tm2, rfa, krum, bylan}.
As result, these aggregators are not a reliable solution for the DL setting where users can connect arbitrarily to each other, and fail in topologies such as a ring.

In order to address this limitation, another category of robust aggregators propose to rely on additional assumptions as compromise~\cite{cao, xie}.
However, these assumptions are typically not realistic (i.e., computing a benign update) or hard to enforce (e.g., access to a clean validation set that generalize across all local datasets).
Hence, these aggregators are not of pragmatic interest, and do not solve the Byzantine problem.

Finally, there is a new but promising trend of robust aggregators in which the user performing the aggregation is only required to trust itself, and which do not impose additional assumptions~\cite{scc, ogclipping}.
Among them, Self Centered Clipping (\SCClip{})~\cite{scc} is showed to perform better against existing attacks~\cite{scc}[Figures 6, 8, 11] and the authors proved its convergence under standard assumptions.
Those points make a strong argument to consider it the state--of--the--art robust aggregator.
We provide more details on \SCClip{} in Section~\ref{sec:scc}.


%% file: parts/3-attacks.tex
\section{Novel robustness attacks}

Existing robustness attacks~\cite{scc, labelflip, backdoorfl, ipm, alie} aim to impair the utility of the the models held by honest users, and cause their accuracy to drop, either on a handful of pre--selected samples, or 
across all samples.
In the context of DL, previous work has use the Byzantine user's knowledge of individual updates to reach these objectives more effectively.
For example, the Dissensus attack~\cite{scc} exploit this knowledge to craft adversarial updates that maximize the distance between honest users, and prevent them to converge to an accurate model.

In this section, we show that (i) this knowledge actually enables a Byzantine user to achieve objectives much more damaging than accuracy drop; and (ii) that the ability to send different updates to users enables the Byzantine user to segregate victims.


\textbf{Threat Model:}
In this work, we use the standard Byzantine threat model~\cite{history, scc}.

Let $\honestUsers$ and $\byzantineUsers$ respectively denote the sets of honest and Byzantine users.
We consider that there is \textbf{one} Byzantine user in the network, i.e., $\byzantineUsers = \{\adv\}$.
By definition, the Byzantine user is active, meaning that they are able to inject {maliciously} crafted updates into the network (line 4 of Algorithm~\ref{alg:scc}).
Under this model, the adversary is assumed to be omniscient, meaning that they know all messages sent on the network, and communication links and their weights~\cite{history, scc}.
In addition, the adversary has knowledge of public training parameters such as the learning rate.
Unlike~\cite{alie, suya}, we do not assume knowledge of local training sets, nor statistical information.

The standard Byzantine threat model assumes that the position of the adversary in the network does not allow it to split the network, i.e., there are no two honest users that communicate only through a Byzantine node. 

During the communication step (2), Byzantine users can deviate from the protocol and craft and send different updates to different neighbors. 
To model these inconsistencies, we add the receiver's index to adversarial updates, i.e., we let ${\theta}_{\adv i}^{t+1/2}$ denote the update sent from $\adv$ to user $i$ at epoch $t$.

\subsection{The state--override attack}
\label{sec:so}

While preventing consensus is a compelling Byzantine objective, the adversary can actually benefit from preserving convergence of the training process, provided that the final model contains adversarial functionalities.
In DL, the adversary can use its access to individual users updates to go beyond injecting perturbations.
Indeed, when knowing the model updates used as input to the aggregation performed by a honest user, the adversary can craft an update that cancels these updates~\cite{privacyDL}, and control the output of the aggregation of its neighbors by adding a payload.
\textbf{As result, the adversary can override the state of honest users, and eventually decide on the exact model the network is going to converge to.}
Using this idea, we propose the \textit{state--override} attack as a robustness attack. 

During the state--override attack, the Byzantine uses its knowledge of the model updates sent in the network to compute the update that cancels out the contributions coming from honest users, and overwrite their local state with a model of its choice. 
Computing the network contribution is easy for the omniscient Byzantine adversary who knows the set of updates $\theta_j^{t+1/2}$ sent by the honest neighbors $j$s of user $i$ (including the update of user $i$ itself). 
By scaling each update using the appropriate entries of $\bm{W}$ and adding them, the Byzantine user eventually recovers the exact contribution from the neighbors of user $i$ for its next state: $\sum_{j \in \honestUsers}\theta_jW_{ij}$ in the case of naive aggregation. 
When negating this value, the adversary obtains a vector that cancels the network contribution.
Finally, by adding a payload, in our case $+\theta_{target}$, to the update, the Byzantine user is able not only to compensate the network contribution, but to replace it by a chosen value.

In DL, users receive a different set of model updates at each epoch, determined by their connections.
As result, the network contribution received by user $i$ is different to the one received by user $j$, and the adversarial update that is computed to override the state of user $i$ will likely fail to have the same effect on user $j$.
Instead, the state--override attack can exploit decentralization by personalizing the adversarial update sent to each of its neighbors according to their network contribution.

Formally, during the state--override attack, the Byzantine user sends:~$ \bm{\theta}_{\adv i} = \frac{\theta_{target} - \sum_{j \in \honestUsers} \theta_jW_{ij}}{W_{ib}}$
to each of its neighboring user $i$.
When naive aggregation is used, it is straightforward that $\theta_i$, the result of the aggregation at user $i$, simplifies to $\theta_{target}$. Indeed:
\begin{align*}
    \theta_i &= \sum_{j \in \honestUsers}\theta_jW_{ij} + \theta_{\adv i}W_{i\adv}\\
    &= \sum_{j \in \honestUsers}\theta_jW_{ij} +\frac{\theta_{target} - \sum_{j \in \honestUsers} \theta_j W_{ij}}{W_{ib}}W_{ib} \\
    &= \sum_{j \in \honestUsers}\theta_jW_{ij} + \theta_{target} - \sum_{j \in \honestUsers} \theta_j W_{ij} \\
    &= \theta_{target}.
\end{align*}
Hence, the state--override attacks allows the Byzantine user to exploit its knowledge of model updates sent across the network and its ability to send inconsistent updates to completely override the neighbor users' state with a target model.

While the state--override attack is instantaneous in the case of naive aggregation, it is likely not the case when robust aggregation is used.
As robust aggregators typically bound the amount of change that are made to locals state at each iteration, it will probably require many more iterations to fully override the neighbors' state, if the attack happens to be feasible. 

\subsection{The sandtrap attack}
\label{sec:st}
Existing robustness attacks aim to attack all honest users at the same time.
However, in DL, there are actually as many outputs $\theta_i^{final}$ as there are users.
Indeed, users have different local states $\theta_i^t$, and, as the gossip--based learning process goes on, users converge towards the final model in their own disparate way.
The fact that users' states conventionally differ during training in DL opens the doors to robustness attacks which aim to affect only one (or few) victim user(s).
To illustrate this new attack concept, we propose the \textit{sandtrap} attack, a \textit{targeted} attack which aims to exclude a victim user from the collaborative training process. 

During a sandtrap attack, the Byzantine user aims to disrupt the learning progress of a targeted user and cause its local state to diverge from the rest of the network.
\textbf{The effect of the sandtrap attack is similar to an attack in which the connections of the victim to the network are removed}: the victim does not benefit from collaboration, and gets trapped at its local state; while the training of the rest of the honest users is not disturbed, beyond the removal of the victim.\looseness=-1

The sandtrap attack can be seen as a targeted state--override attack, where the adversary uses its knowledge of individual model updates to send to the victim an update that will cancel out the contributions of the victim's neighbors.

To non--targets, the adversary sends an update that ensures that their learning process is not hindered.
A good candidate for this update is simply the average of all updates produced by honest users, excluding the one coming from the victim.

More formally, the adversary sends the following update to the victim $\victim$:~$ \bm{\theta}_{\adv \victim} = \frac{\theta_{\victim} - \sum_{j \in \honestUsers} \theta_jW_{\victim j}}{W_{\victim b}}$, 
where $\theta_{\victim}$ is the update of the victim, and $- \sum_{j \in \honestUsers} \theta_jW_{\victim j}$ is the negation of the updates coming from the neighbors of the victim.
Thus, when used by the victim in the aggregation step, this update cancels out the contributions of the honest updates received by the victim, which leaves the victim learning only from its own update~$\theta_{\victim}$.
The victim is then ``sandtrapped" to its own local state.
When naive aggregation is used, we can see that the victim computes its next state as:
\begin{align*}
    \theta_\victim &= \sum_{j \in \honestUsers}\theta_jW_{\victim j} +\theta_{\victim} - \sum_{j \in \honestUsers} \theta_j\frac{W_{\victim j}}{W_{\victim b}}W_{\victim b} 
    = \theta_\victim.
\end{align*}
On the other hand, the adversary sends to non--target users:~$ \bm{\theta}_{\adv i} = \sum_{j \in \honestUsers \setminus \{\victim\}} \theta_j W_{i j}.$

%% file: parts/4-experiments.tex
\section{Experiments and Results}
\label{sec:attacks}

Previously, we presented the (theoretical) outcome of our attacks under naive aggregation.
In this Section, we empirically evaluate the effectiveness of our attacks against a robust deployment of DL.
As robust aggregator, we use the state--of--the--art aggregator \SCClip{}~\cite{scc}.

\subsection{Robust DL via Self Centered Clipping}
\label{sec:scc}

The \SCClip{} aggregator is implemented by each user $i$ at step (3), i.e., line 6 of Algorithm \ref{alg:scc} as follows:
\begin{align*}
    \SCClip_i(\theta_1^{t+1/2}, ..., \theta_n^{t+1/2}) :&= \sum_{j=1}^n W_{ij}(\theta_i^{t+1/2} + \\
    &\Clip(\theta_j^{t+1/2}-\theta_i^{t+1/2}, \tau_i^{t})),
\end{align*} 
where $\tau_i^t$ is an additional hyper--parameter called the clipping radius and $\Clip(z, \tau) := \texttt{min}(1, \frac{\tau}{||z||})\cdot z$.
Intuitively, \SCClip{} bounds the model updates received by the user $i$ that differ from $\theta_i$ by $\tau_i$.
Each user uses its local state as a trusted reference point.
The difference of the received updates with the reference is clipped by a radius $\tau$, in order to only make bounded steps away from the trusted point at each iteration.
The progress of a user to the final model is then dominated by its own updates.

\textbf{Clipping radius.}
The choice of $\tau$ is pivotal. If the radius is too big, there is less robustness as the adversaries are given more influence \cite{scc}. If too small, more epochs are needed to reach consensus, which defeats the initial purpose of DL---namely efficiency---in addition to threatening the privacy of users~\cite{privacyDL}.
Several ways to select $\tau_i^{t}$ are described:

Ideally, $\tau_i^t$ is computed as the average variance of honest users around user $i$, i.e.:
\begin{equation}
\label{eq:tau}
    \tau_i^{t} := \sqrt{\frac{1}{\delta_i} \sum_{j \in \honestUsers}W_{ij}||\theta_i^{t+1/2}-\theta_j^{t+1/2}||_2^2},
\end{equation}
where $\delta_i$ is the total weight of Byzantine edges around user~$i$.
However, as highlighted in~\cite{scc}, Eq.~\ref{eq:tau} cannot be computed in practice due to the subset of honest users (or equivalently the subset of Byzantine users) being unknown.
Instead, $\tau_i^{t}$ can be a pre--determined and constant value.
Following the evaluation setup of~\cite{scc}, we set this value to~$\tau_i^{t} := 1$.

\subsection{Graphs and dataset} For our experiments, we use the MNIST dataset \cite{mnist}, split among users in a non--IID way to capture a more realistic setting.\footnote{Data is partitioned by class among the pool of users.}
We train a model with a shallow CNN architecture: layers = Conv-Conv-Dropout-fc-Dropout-fc; and using default training parameters: batch size = 32, momentum $\alpha$ = 0.9, learning rate $\eta$ = 0.01 without decay.

We consider 9 users, 8 honest and one Byzantine.
We use two different topologies to connect them: the torus (Figure~\ref{fig:torus}) and the dumbbell topologies (Figure~\ref{fig:dumbbell}).
The torus is quite popular in the dencentralized community, and the dumbbell is a topology in which \SCClip{} should excel~\cite{scc}.
We assign equal weights to edges, i.e.,~$\bm{W}_{ij} = \frac{1}{|\mathcal{N}_i|}.$

\input{images/topo.tex}

All assumptions on the graph and parameters of \SCClip{} are fulfilled, which means that our evaluation setup is fair.

\subsection{Results}

\input{images/so_dist.tex}
\input{images/atk_acc.tex}

\subsubsection{State--override attack}
As described in Section~\ref{sec:so}, the objective of the state--override attack is to replace the set of local parameters of the honest users with an arbitrarily chosen target model~$\theta_{target}$.
Thus, a natural success metric for the attack is, for any targeted honest user~$i$, the distance from $i$'s current parameters and the adversary's target model; formally:~\mbox{$D_i := ||\theta_i - \theta_{target}||^2.$}

For simplicity, in our experiments we choose the target model to be the \textbf{\textit{all--0}} model, i.e., $\theta_{target} = \{0\}^d$ with $d$ the model size. 

We report the evolution of $D_i$ for all users across epochs in Figure~\ref{fig:so} for both variants of the clipping radius---ideal (Eq.~\ref{eq:tau}), in blue and constant (i.e., $\tau_i^{t} := 1$), in orange.
We run the attack in the torus (Figure~\ref{fig:so_t}) and the dumbbell topology (Figure~\ref{fig:so_d}).
We also report in Table~\ref{tab:acc} the accuracy of honest users' model at the end of the training process, i.e., after~$300$ epochs.
The variance is below 0.5\%, so we omit it from the table and only report the average accuracy of users' model.

In few epochs, the states of the neighbors of the adversary are overwritten, and by half of the training process,
all users in the network converge to the target state.
Indeed, once the state of the adversary's neighbors is overwritten, they unwillingly propagate the perturbation to their own neighbors in a self--feeding loop.
For example for the scenario of DL with \SCClip{} using ideal clipping radius in the dumbbell topology, the distance between the models of the adversary's neighbors and the target model drops from 82 to 1 in 30 epochs, and by epoch 60, the distance between each users' model and the target goes below $0.01$.
The number of epochs needed to overwrite the state of \textit{all} users depends on the topology and clipping radius, but is at most 150 epochs.
This number is relatively low compared to usual total number of epochs used in the collaborative setting ($1e3$ to $1e6$ in~\cite{scc}).
As consequence on the utility, the accuracy of honest users' models drops very low in all experiments, 9.8\% to be precise, as the all--0 model produces random predictions.

\textbf{On the impact of the clipping radius.}
We notice that the state--override attack requires less epochs to overwrite users' state when the \textit{ideal} clipping radius (blue line) is used in comparison to the constant alternative (orange line).
We find out that when the ideal clipping radius is used, the adversary is able to manipulate and increase the value of $\tau$ through the state--override attack.
As $\tau$ gets larger, the adversary is able to introduce more change on the state of the honest users, hence overwriting the users' state in less epochs.
\textbf{This observation comes with the finding of a new attack surface:~$\tau$ manipulation.}
We expand on this in Appendix~\ref{sec:tau_atk}.

\input{images/st_acc.tex}
\textbf{On the impact of the topology.}
When comparing the two graphs of the results of the state--override attack, we see that the state of the honest users is overwritten by the target model faster under the torus topology (Figure~\ref{fig:so_t}), compared to the dumbbell (Figure ~\ref{fig:so_d}).
The reason behind this difference is the connectivity of the adversary (4 neighbors in the torus, but only 2 in the dumbbell) but also the connectivity of the adversary's neighbors. In the torus topology, the neighbors of the Byzantine user have less neighbors (4) than in the dumbbell one (5). 
As the adversary has less influence on its neighbors and can reach fewer honest users directly in the dumbbell topology, it takes longer to overwrite the state of all users in the network. 
However, this difference is not significant and we can safely state that the attack works well in both topologies, for any choice of clipping radius.

\subsubsection{Sandtrap attack.}
Next, we look at the outcome of the sandtrap attack.
In our experiments, the victim $\victim$ is picked as the first user among the adversary's list of neighbors, i.e., user $u_3$. 
We report the accuracy of the victim's model and the average accuracy of non--victims' models across epochs in Figures~\ref{fig:st_t} and~\ref{fig:st_d}, and at the end of the training in Table~\ref{tab:acc}.

We can see that the sandtrap attack causes the victim to hold a non--accurate model (with accuracy from $23.5\%$ down to $9.2\%$), while the rest of the honest users obtain models relatively less impacted with regards to accuracy ($62.1\%$ to $35.3\%$), especially if we consider the baseline accuracy of models resulting from a training process where the victim is excluded.

\textbf{On the impact of the topology.} We observe that the average accuracy of non--target users is lower in the dumbbell topology (Figure~\ref{fig:st_d}) than in the torus topology (Figure~\ref{fig:st_t}). 
In the case of the dumbbell topology, the victim acts as a bridge between two groups: the first group (G1) with users $u_1$, $u_2$, and $u_4$, and the second group (G2) with users $u_5$, $u_6$, $u_7$ and $u_8$.
As consequence, when excluding the victim from the network by performing the sandtrap attack, the victim stops reliably relaying information from its neighbors and model updates do not satisfactorily circulate in--between the two groups.
Then, users from G1 and G2 only learn from the users in their respective group, which causes their performance to appear very low compared to the baseline which considers the learning protocol with contributions from the whole network.
Furthermore, in addition to being deprived from contributions by the other half of the network, the users from G1 will unlikely filter all contributions of poor quality coming from the victim, lowering their accuracy even more.
For completeness, we report the detailed accuracy of both groups G1 and G2 in Table~\ref{tab:acc}, and see that the accuracy of users in G1 is lower than the accuracy of users in G2.\looseness=-1
%

%% file: images/topo.tex
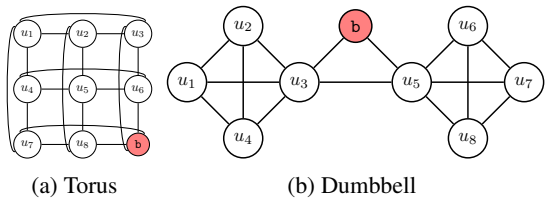
\begin{figure}[t]
\centering
\subfloat[Torus]{\label{fig:torus}
\resizebox{.12\textwidth}{!}{
\begin{tikzpicture}[node distance={15mm}, thick, main/.style = {draw, circle}] 
\node[main] (0) {$u_1$}; 
\node[main] (1) [right of=0] {$u_2$}; 
\node[main] (2) [right of=1] {$u_3$}; 
\node[main] (3) [below of=0] {$u_4$}; 
\node[main] (4) [below of=1] {$u_5$}; 
\node[main] (5) [below of=2] {$u_6$}; 
\node[main] (6) [below of=3] {$u_7$}; 
\node[main] (7) [below of=4] {$u_8$}; 
\node[main, fill=red!50] (a) [below of=5] {$\adv$}; 

\draw (0) -- (1); 
\draw (0) -- (3); 
\draw (1) -- (2); 
\draw (1) -- (4); 
\draw (2) -- (5); 
\draw (3) -- (6); 
\draw (3) -- (4); 
\draw (7) -- (a); 
\draw (4) -- (5); 
\draw (4) -- (7); 
\draw (5) -- (a); 
\draw (6) -- (7); 
\draw (0) to [out=120, in=60, looseness=0.25] (2);
\draw (3) to [out=120, in=60, looseness=0.25] (5);
\draw (6) to [out=120, in=60, looseness=0.25] (a);
\draw (0) to [out=150, in=210, looseness=0.25] (6);
\draw (1) to [out=150, in=210, looseness=0.25] (7);
\draw (2) to [out=150, in=210, looseness=0.25] (a);

\end{tikzpicture}


}}\subfloat[Dumbbell]{\label{fig:dumbbell}
\resizebox{.3\textwidth}{!}{

\begin{tikzpicture}[node distance={15mm}, thick, main/.style = {draw, circle}] 
\node[main] (0) {$u_1$}; 
\node[main] (1) [above right of=0] {$u_2$}; 
\node[main] (2) [below right of=0] {$u_4$}; 
\node[main] (3) [above right of=2] {$u_3$}; 
\node[main, fill=red!50] (a) [above right of=3] {$\adv$}; 
\node[main] (4) [below right of=a] {$u_5$}; 
\node[main] (5) [above right of=4] {$u_6$}; 
\node[main] (6) [below right of=4] {$u_8$}; 
\node[main] (7) [above right of=6] {$u_7$}; 

\draw (0) -- (1); 
\draw (0) -- (2); 
\draw (0) -- (3); 
\draw (1) -- (2); 
\draw (1) -- (3); 
\draw (3) -- (2); 
\draw (3) -- (4); 
\draw (3) -- (a); 
\draw (4) -- (5); 
\draw (4) -- (6); 
\draw (4) -- (7); 
\draw (4) -- (a); 
\draw (5) -- (7); 
\draw (5) -- (6); 
\draw (6) -- (7); 
\end{tikzpicture} 
}}
\caption{The torus and dumbbell topology with 8 honest users and 1 adversary.}
\end{figure}

%% file: images/so_dist.tex
\begin{figure*}[ht!]
    \centering
    \hspace{1.5cm}  \includegraphics[width=0.25\linewidth, trim = 0mm 2mm 0mm 0mm, clip]{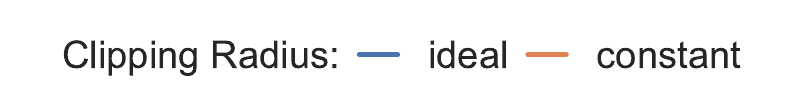} \hspace{3.5cm}     \includegraphics[width=0.41\linewidth, trim = 0mm 2mm 0mm 0mm, clip]{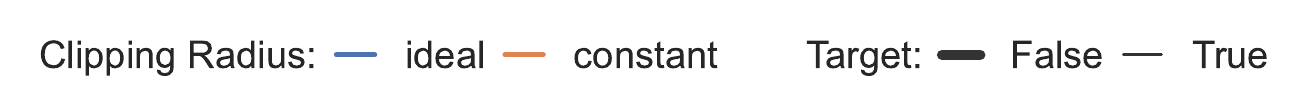}
    \subfloat[\scriptsize{Torus (state-override)}]{\label{fig:so_t}
        \includegraphics[width=0.23\linewidth]{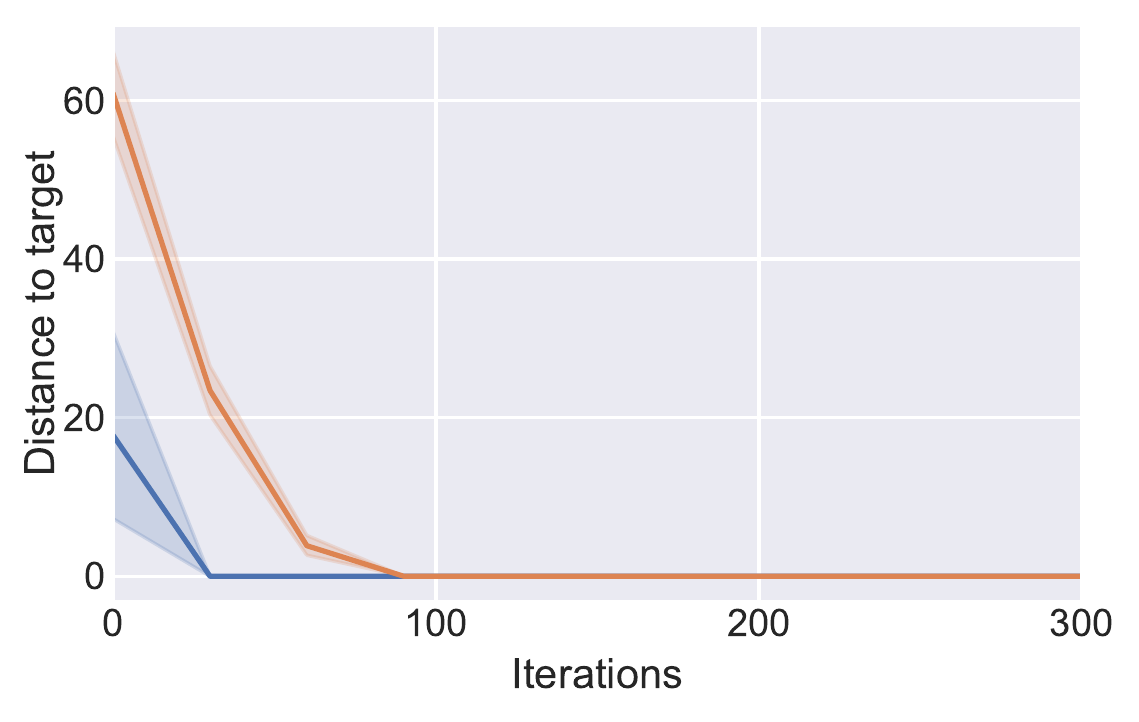}
   } \subfloat[\scriptsize{Dumbbell (state-override)}]{\label{fig:so_d}
        \includegraphics[width=0.23\linewidth]{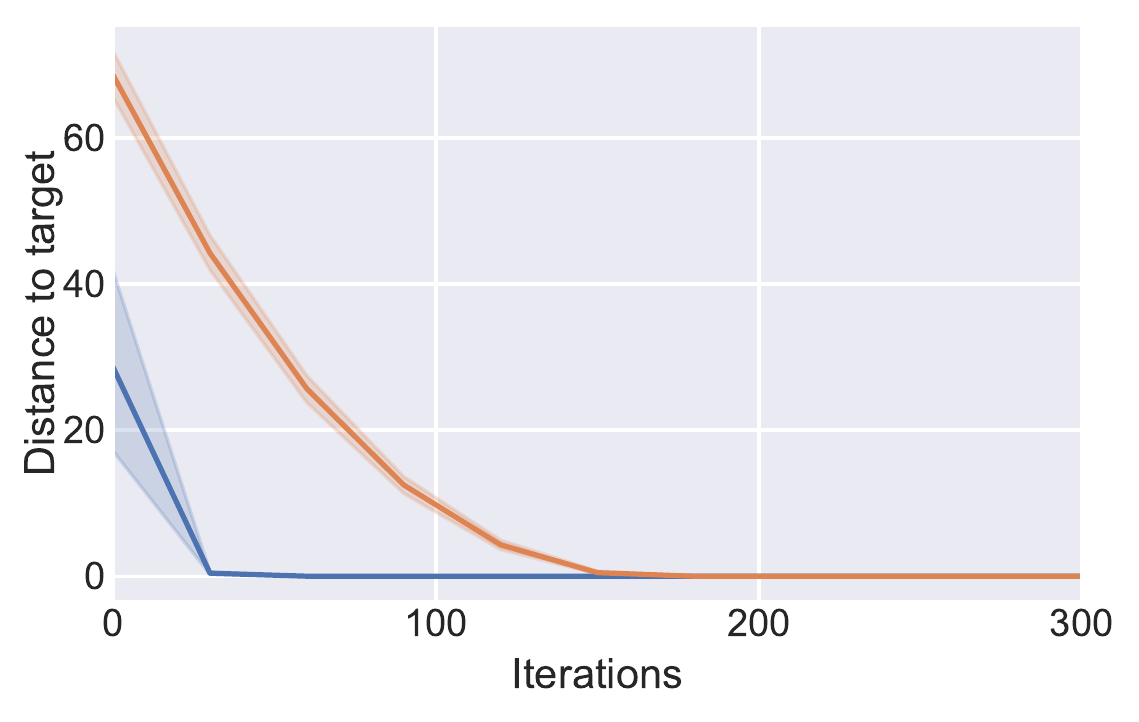}
    } \hspace{.5cm} \subfloat[\scriptsize{Torus (sandtrap)}]{\label{fig:st_t}
        \includegraphics[width=0.23\linewidth]{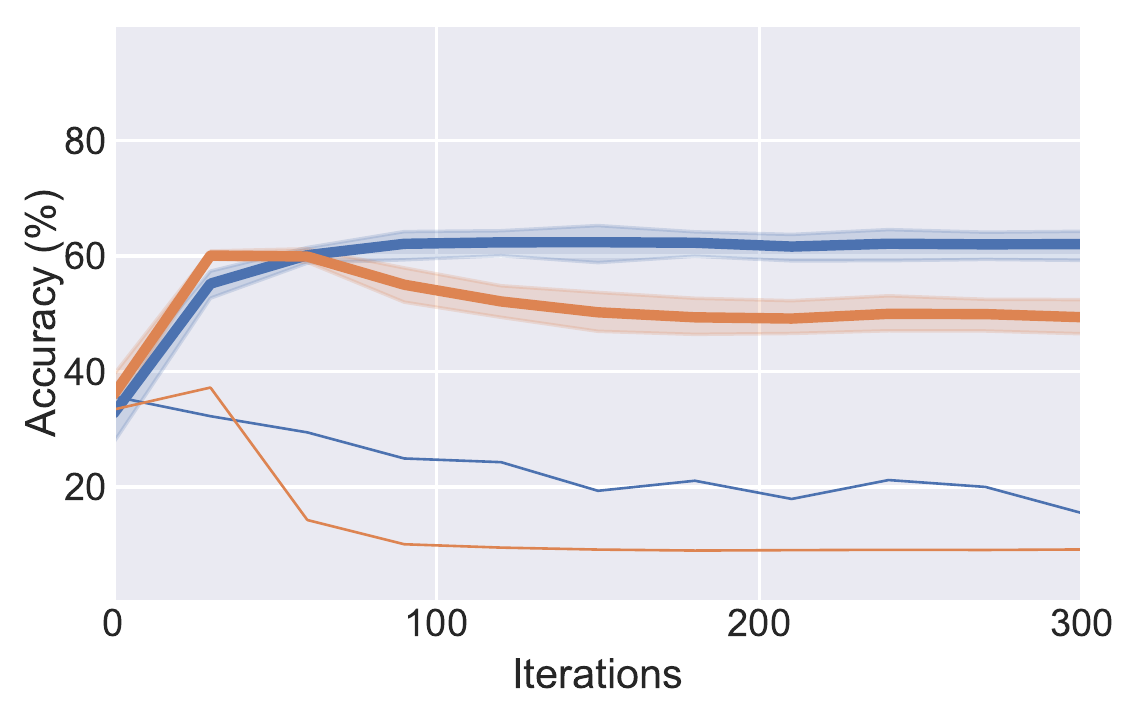}
   } \subfloat[\scriptsize{Dumbbell (sandtrap)}]{\label{fig:st_d}
        \includegraphics[width=0.23\linewidth]{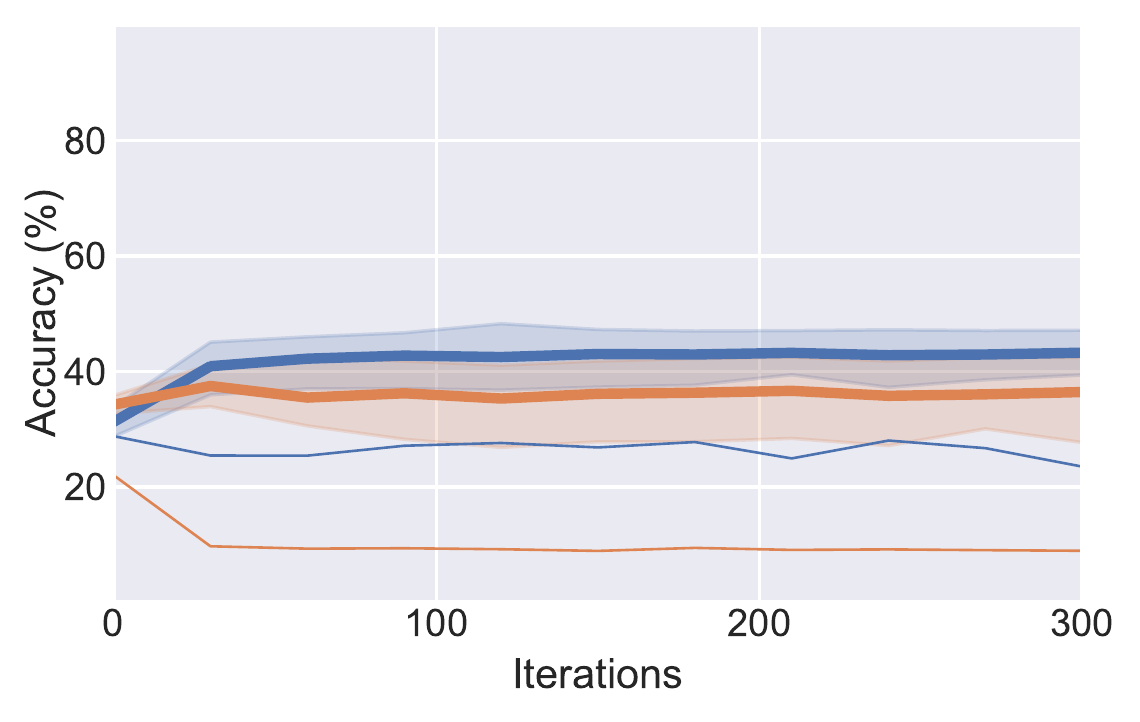}
    }
    \caption{Panels~(a) and (b):~distance to target under the state--override attack. Panels~(c) and (d):~accuracy of the target's model versus non-targets' models under the sandtrap attack. The shaded regions represent the variance accross users.}
    \label{fig:so}
\end{figure*}

%% file: images/atk_acc.tex
\begin{table}[t]
\caption{Average classification accuracies of users' local model after 300 epochs. ``T'' is short for torus and ``D'' for dumbbell. }
\label{tab:acc}
\begin{center}
\begin{scriptsize}
\begin{sc}
\begin{tabular}{llrr}
\toprule
& & ideal $\tau$ & constant $\tau$\\
\midrule
\multirow{4}{*}{T} & baseline & 83.9\% & 87.0\% \\
                    & state--override & 9.8\% & 9.8\% \\
                    & sandtrap, target & 15.4\% & 9.2\% \\
                    & sandtrap, non-targets & 62.1\% & 49.4\% \\
\midrule   
 \multirow{6}{*}{D} & baseline & 62.6\% & 80.8\%\\
                    & state--override & 9.8\% & 9.8\% \\
                    & sandtrap, target & 23.5\% & 9.0\%\\
                    & sandtrap, non-targets & 43.3\% & 35.3\% \\ 
                    & \qquad $\hookrightarrow$ G1 & 35.8\% & 24.7\% \\ 
                    & \qquad $\hookrightarrow$ G2 & 48.8\% & 45.2\% \\ 
\bottomrule
\end{tabular}
\end{sc}
\end{scriptsize}
\end{center}
\vskip -0.2in
\end{table}

%% file: images/st_acc.tex
%% file: parts/5-DLvsFL.tex
\section{Decentralization and robustness}
\label{sec:comparison}

In this section we compare DL to Federated Learning (FL), another CML approach that relies on a central server to orchestrate the training.
Advocates of DL assert that DL is more robust than FL~\cite{splitfed, scc, ogclipping, survey}.

\noindent\textbf{Federated Learning Background. }
In Federated Learning (FL), the training process is coordinated by a server and there is no communication between users.
The server occupies a central position and is connected to all users, in a star topology (see Figure~\ref{fig:fl}).
To perform the training, the server decides on an initial set of global parameters~$\theta^0$, and the following steps are repeated until a stop condition is met: (1) \textit{Local training:} Users apply gradient descent on the global model parameters~$\theta^{t}$. 
The output of the gradient descend is referred to as the \textit{model update}~$\theta_{i}^{t+1/2}$.
(2) \textit{Communication:} Users send their model update $\theta_{i}^{t+1/2}$ to the server.
(3) \textit{Aggregation:} Upon receiving all updates, the server aggregates them (e.g. average of the users' updates), use the result of the aggregation to update the current global model, $\theta^{t+1}$, and broadcasts this model back to the users.

\input{parts/5a-influence.tex}

\input{parts/5c-locality.tex}
\input{parts/5b-inconsistencies.tex}
\input{parts/5d-knowledge.tex}

%% file: parts/5a-influence.tex
\subsection{Adversarial influence on victim's model}
\label{sec:influence}
In FL, at each round, the server aggregates contributions from \textit{all} users to compute the next global model. 
All users contribute to the learning progress equally. 
Under the assumption of bounded contributions---which can be trivially enforced by the server even when secure aggregation is used~\cite{acorn}---users have influence of $1/n$ on the global model.

On the contrary, in DL, each user learns from and aggregates the updates of only a subset of users: its neighbors. 
Therefore, at each epoch, each user contributes to the learning process of its neighbors only.
The influence of a user on the model of its neighbor $i$ is determined by how many users send updates to this user $i$, i.e., its number of neighbors $\mathcal{N}_i$.
Hence, users in DL have influence of $1/|\mathcal{N}_i|$ on the model of their neighbor $i$.

\begin{figure}[t!]
    \centering
    \subfloat[]{\label{fig:conn}
    \includegraphics[width=0.25\textwidth]{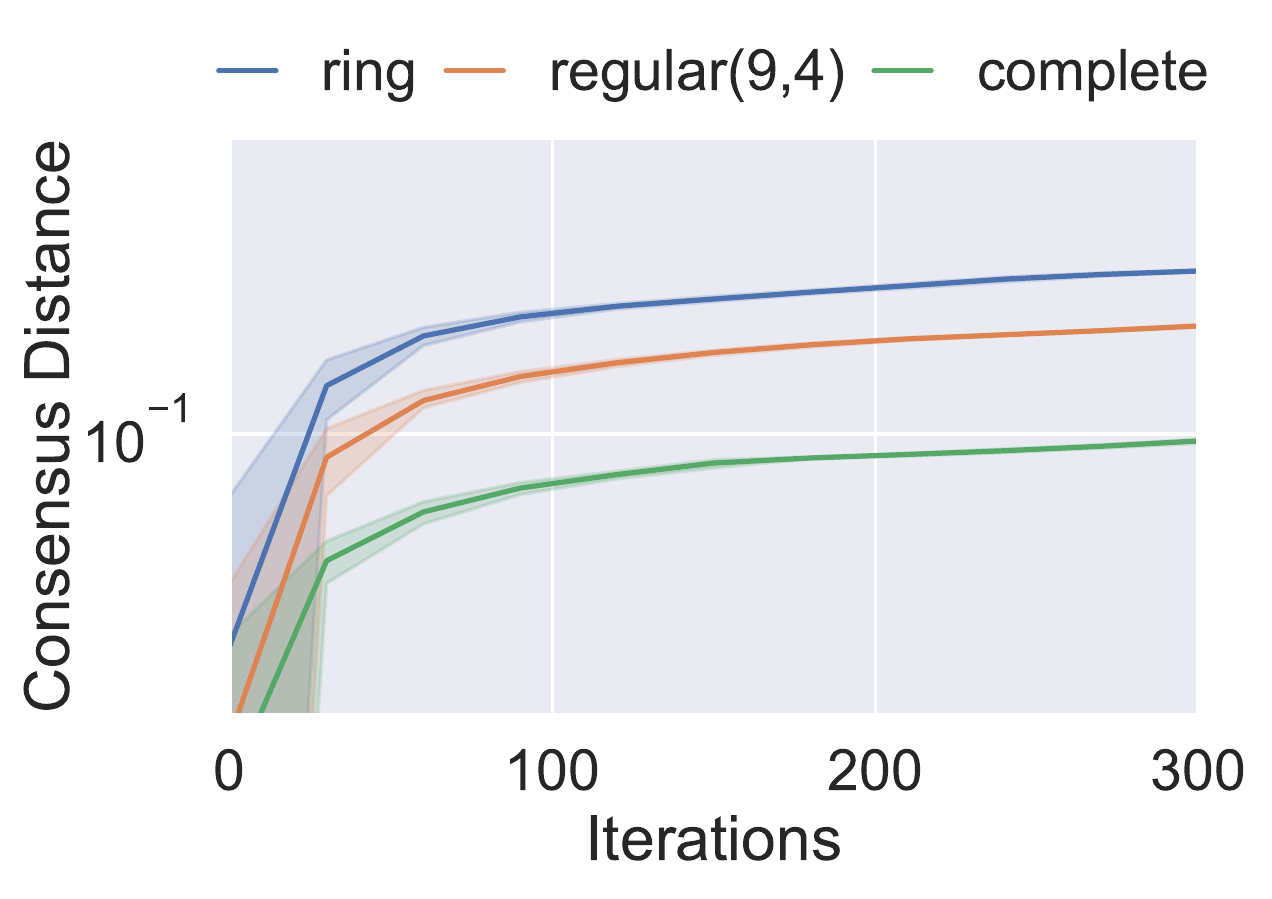}
    }
    \subfloat[]{\label{fig:inconst}
    \includegraphics[width=0.25\textwidth]{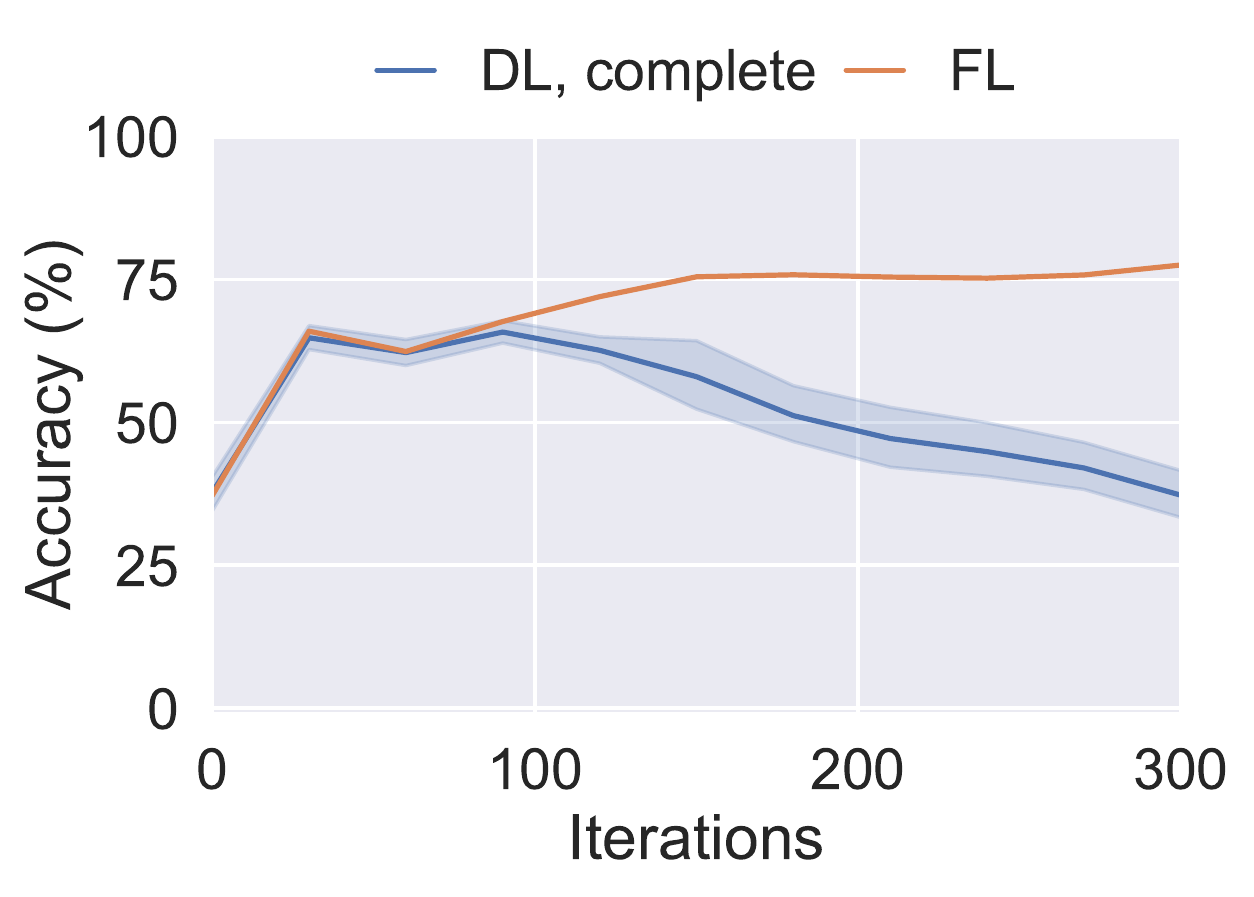}
    }
   
    \caption{ Panel~(a): consensus distance in DL under the dissensus attack using increasingly connected topologies. Panel~(b): accuracy of users' models in DL with complete topology and of the global model in FL, under the noisy attack.}
\end{figure}

Influence on the input translates in influence on the output of the (robust) aggregation.
The larger influence the adversary has on the output of the aggregation, the more it can tamper with the resulting model.
In DL, for efficiency purposes, the topology is typically spare, meaning that the number of neighbors $|\mathcal{N}_i|$ of each user $i$ is significantly smaller than the total number of users $n$ ($|\mathcal{N}_i| << n$, $1/|\mathcal{N}_i|$).
Thus, the influence of Byzantine users in DL is higher than in FL.
As a result, an adversary in DL has more control on the state of honest users compared to FL. 
\textbf{In sparse topologies, decentralization decreases the robustness of the system compared to FL by increasing the adversarial influence of Byzantine users on honest users' states.}

Since the robustness is a function of  $|\mathcal{N}_i|$, it is possible to increase robustness by increasing the connectivity of users in DL.
As $|\mathcal{N}_i|$ increases, $1/|\mathcal{N}_i|$ decreases and the adversarial update gets aggregated with more honestly generated model updates.
Robustness is maximized when users are fully connected ($|\mathcal{N}_i|=n$).

We empirically evaluate the relation between connectivity and robustness in DL.
We measure the effectiveness of a Byzantine attack in increasingly connected topologies.
During the experiment, only the connectivity of the honest users is changing, and the number of neighbors of the adversary stays constant.
This allows us to isolate the effect of connectivity in robustness.
We consider 9 decentralized users, 8 honest and 1 Byzantine, using naive aggregation.
In each topology, the Byzantine user performs a dissensus attack~\cite{scc}.
We first connect the participants using a ring topology (``$ring$'').
We progressively add edges in--between honest users to form a regular graph where each honest user is connected to 4 other honest users (``$regular(9, 4)$'').
Eventually, the honest users become fully connected (``$complete$''). 
As the objective of the dissensus attack is to push users away from each other, the best metric to measure the effectiveness of the dissensus attack is the distance of users' set of parameters to the average.
This is captured by the consensus distance, computed as: $C_i := || \theta_i - \frac{1}{|\honestUsers|}\sum_{j \in \honestUsers}\theta_j||^2$. 
The higher the consensus distance, the more effective the attack.
We report our results in Figure~\ref{fig:conn}.
We observe that the complete topology offers the most robustness (lowest consensus distance), while the ring---the topology where users only have 2 neighbors---is the most vulnerable topology to the dissensus attack as it has the highest consensus distance among users.
These results confirm that \textbf{increasing the connectivity of users increases the robustness of DL.}

In DL, robustness is maximized under the complete topology, when the adversarial influence is reduced to~$1/n$.
Thus, \textbf{the robustness provided by DL can at best match the robustness provided by FL}.
However, as we show in Section~\ref{sec:incons}, this upper--bound is strict: even when the robustness in DL is maximized---the topology is complete---, it is still strictly less than what is obtained in FL.

We note that, when the topology is complete, DL training is equivalent to FL (the output models are an aggregation of \emph{all} users in the system), while being less efficient. In DL the asymptotic communication cost becomes $O(n^2)$, in comparison with $O(n)$ in FL\footnote{To be precise, $O(n)$ is the communication cost of the busiest participant in FL: the server, while it is only $O(1)$ for the users.}.


%% file: parts/5c-locality.tex
\subsection{Attacking a global model versus local models}
\label{sec:local}

In DL, each user holds a local set of parameters and advances towards the final model disparately.
The training typically stops when consensus among users is met.
In FL, all users share a global set of parameters and improve it simultaneously in each training round.

In FL, any adversarial update is sent to the server, aggregated, and affects all users through the global model users receive from the server.
Hence, a Byzantine user can only perform attacks against all users at the same time, i.e., that affect the global model. 
In DL however, a Byzantine user can use the fact that it is expected that users each have a different local state, to target not the global model but the states of a subset of users only.

Then, \textbf{DL provides Byzantine users with an additional attack capability compared to FL: the possibility to perform targeted attacks on a subset of users.} 
The \textit{sandtrap} attack is an example of a targeted attack in DL where only one user is affected.
It is however impossible for an adversarial federated user to cause the users' states to differ as the update sent to the server affects all users trough the aggregated result computed by the server~\cite{FLpoisonBoard}. \looseness=-1

Based on their own connectivity---number of neighbors---and the connectivity of the adversary, some users might be more vulnerable to robustness attacks than others.
The lower their number of neighbors, the higher the influence of an adversarial neighbor on their local state (see Section~\ref{sec:influence}).
When targeted attacks are enabled, \textbf{the training process in DL is only as robust as its least protected user.}


%% file: parts/5b-inconsistencies.tex
\subsection{Sending one update versus many}
\label{sec:incons}

Any attack that can be implemented by a Byzantine user in FL can be implemented in DL, while the opposite is not true.
In the decentralized setting, a Byzantine user can send different updates to different users, creating what is called \textit{update inconsistencies}~\cite{inconsistencies}.
In FL, the adversary can only send one update---to the server, hence does not have the ability to send inconsistent updates.

We illustrate the impact of the ability to send different updates to different users by comparing the outcome of the same attack (i)~with and (ii)~without updates inconsistencies.
The attack without inconsistencies is performed in FL.
The attack with inconsistencies is obviously performed in DL (the only design where it is possible).
To remove other factors that could influence the results, we use a complete topology in DL. 
Indeed in this case, the adversarial influence is the same ($1/n$) as in FL.
We consider 9 users, 8 honest and 1 Byzantine, and use naive aggregation.
The dissensus attack~\cite{scc} is the only attack that makes use of updates inconsistencies, but its adversarial objective does not apply to FL (the consensus distance is always 0 in FL since there is only one global model).
We then design a toy attack for the purpose of the experiment, that we call the ``noisy attack''.
While simple, this attack enables us to showcase the impact on robustness that comes with the ability to send inconsistent updates.
The noisy attack consists in echoing back a noisy version of the update(s) received.

In DL, the adversary sends to user $i$: \\
\mbox{$\theta_{\adv i}^{t+1/2} = (1-\epsilon)\theta_i^{t+1/2} + \epsilon z$}, 
where $z$ is a random vector uniformly sampled in $[0, 0.01]$ (as weights are typically small) and $\epsilon$ is a hyper--parameter that controls the strength of the attack.
Since we use a complete topology, the Byzantine user can send a different update to each user $i$, i.e., a total of $n$ different updates.

In FL, the adversary can send only \textit{one} update, and cannot make use of update inconsistencies.
In order to match the concept of the noisy attack---the adversary echoes the received updates---, the adversary in FL send a noisy version of the average of all honest users' updates, i.e., $\frac{1}{|\honestUsers|}\sum_{i \in \honestUsers}\theta_i^{t+1/2}$.
In FL, the adversary sends to the server: $\theta_{\adv}^{t+1/2} = (1-\epsilon)\frac{1}{|\honestUsers|}\sum_{i \in \honestUsers}\theta_i^{t+1/2} + \epsilon z$.

We plot the outcome of the noisy attack for $\epsilon=0.05$---small amount of noise---in Figure~\ref{fig:inconst}.
We observe that the accuracy of the users' models in the complete topology in DL drops below the accuracy of the global model in FL.
Hence, the noisy attack performs better against DL than against FL.
Since the functionality of DL in a complete topology and FL is the same, this difference in robustness comes only from the fact that the adversary is able to send different updates to different neighbors.


\textbf{In all topologies,
decentralization decreases the robustness of the system
compared to FL by giving a Byzantine user the ability to send inconsistent updates.}

To be noted that when robust aggregation is used, it is not sufficient to close the robustness gap between the decentralized and federated approaches, as we discuss in Appendix~\ref{sec:ra}.

%% file: parts/5d-knowledge.tex
\subsection{Observing a global update vs. individual updates}

Existing Byzantine attacks (including ours) assume an omniscient adversary and rely on the adversary's knowledge, e.g., model updates sent on the network~\cite{scc, alie, ipm}, edges' weight, or even honest users' training data~\cite{history}.
However, this threat model is very unlikely to happen in practice.

In a typical collaborative learning deployment~\cite{disco, tffl}, users, including Byzantine users, only access local information~\cite{threatmodel_network, splitfed}. They can observe the model updates that they receive from a fixed subset of participants---their neighbors in DL, the server in FL.
In this threat model, the Byzantine federated user observes only one aggregated answer coming from the server.
In comparison, in DL, a Byzantine user observes more model updates with each additional user it connects to.
Furthermore, algorithms in the DL literature allow users to select their neighbors arbitrarily.
An adversary in DL can thus connect to all users in the network, and observe all model updates,
enabling attacks such as the state--override\footnote{The state--override attack is possible on user $i$ as soon as the adversary knows all updates this user receives, i.e., $\mathcal{N}_i \subset \mathcal{N}_\adv$.}.

\textbf{In realistic deployments of CML, a Byzantine decentralized user has access to individual model updates, which are not available in FL.
Knowledge of these individual updates enables powerful attacks (e.g., state--override).}

%% file: parts/6-discussion.tex
    \section{Takeaways}
\label{sec:conclusion}

In this work, we introduce two robustness attacks against decentralized learning protocols. 
These novel attacks show that: 
(i) ~Byzantine users in DL have more capabilities than previously considered in the literature,
(ii)~Byzantine users can determine the model resulting from the collaborative learning process, and can isolate victims without tampering with their connections
(iii)~state--of--the--art robust aggregation techniques fail to offer the promised level of robustness, 

We also show that,
due to the extra capabilities that peer-to-peer connections provides to Byzantine users compared to communications being proxied by a centralized server, DL cannot provide more robustness against Byzantine users than FL. 
Finally, we show that, in DL, robustness and topology sparsity are in direct opposition. This suggests that achieving robustness while preserving communication efficiency may be impossible.

%% file: app/appendix.tex
\section{Extra material}
\label{sec:dl}

\begin{algorithm}[h]
    \caption{DL}
    \label{alg:scc}
    \begin{algorithmic}[1]
    \REQUIRE $\bm{\theta}^0 \in \mathbb{R}^d, \alpha, \eta, \bm{m}_i^0 = \bm{f}_i(\bm{\theta}^0; X_i)$
    \FOR{$t = 0, 1, ...$}
        \FORALL{$i = 1, ..., n$}
            \STATE $\bm{m}_i^{t+1} = (1-\alpha)\bm{m}_i^{t}+\alpha \bm{f}_i(\bm{\theta}_i^t; X_i)$
            \STATE $\bm{\theta}_i^{t+1/2} = \bm{\theta}_i^t - \eta \bm{m}_i^{t+1}$
            \STATE Exchange $\bm{\theta}_i^{t+1/2}$ with $\mathcal{N}_i$.
            \STATE $\bm{\theta}_i^{t+1} =$ \texttt{Aggregate}$(\bm{\theta}_1^{t+1/2}, ..., \bm{\theta}_n^{t+1/2})$
        \ENDFOR
    \ENDFOR
    \end{algorithmic}
\end{algorithm}

\input{images/cl_schema.tex}

\section{On the impact of the clipping radius}
\label{sec:tau_atk}
In Figure~\ref{fig:so}, we notice that the state--override attacks works better when the ideal clipping radius is used, compared to when the clipping radius is constant.
To understand why, we plot in Figure~\ref{fig:tau} the evolution of the clipping radius during the state--override attack and during a benign training process (``\textit{baseline}'').
At the beginning of the training process, the value of $\tau$ under attack is significantly higher than the value of $\tau$ in the benign case (up to $\times 1000$).
We derive that the adversary (indirectly) influences the value of the clipping radius when Eq.~\ref{eq:tau} is used.
When compensating for the network contribution ($-\sum_{j \in \honestUsers}\theta_j$), the adversary augments the distance between users' states, hence the value of $\tau$.
As the clipping radius gets larger, the adversary can cause larger perturbation on the honest users' local state.
In comparison, when the clipping radius is fixed to 1, it requires more epochs to cause the same amount of perturbation, explaining the difference. 
While this phenomenon happened ``accidentally'' with the state--override attack, it can be abused by a Byzantine attacker which purposefully increases the value of the clipping radius, e.g., by introducing dissensus into the network~\cite{scc}, to be able to send poisoned updates with large perturbations later on.
Better convergence through the use of ideal clipping radius comes at the cost of increasing the attack surface.
\input{images/so_tau.tex}

\input{parts/5e-aggregators.tex}

%% file: images/cl_schema.tex
\begin{figure}[h]
    \centering
    \subfloat[Federated Learning]{\label{fig:fl}
        \resizebox{0.2\textwidth}{!}{%
		\begin{tikzpicture}
			\tikzstyle{user} = [fill=green!30, circle,  text width=5mm, align=center]
			\node[user, fill=yellow!30] (1) {S}; 
			
			\node[user, yshift=-2.5cm, xshift=-1.8cm] (2) {$u_1$}; 
			\node[user, yshift=-2.5cm, xshift=-0.6cm] (3) {$u_2$}; 
			\node[user, yshift=-2.5cm, xshift=+0.6cm] (4) {$u_3$}; 
			\node[user, yshift=-2.5cm, xshift=+1.8cm] (5) {$u_4$}; 
				
			\draw[<->] (2) --  (1);
			\draw[<->] (3) -- (1);
			\draw[<->] (4) -- (1);
			\draw[<->] (5) -- (1);
		\end{tikzpicture} 
	}
    }
    \subfloat[Decentralized Learning]{\label{fig:dl}
    \resizebox{0.2\textwidth}{!}{%
			\begin{tikzpicture}
		
			\tikzstyle{user} = [fill=green!30, circle,  text width=5mm, align=center]
			\node[user] (1) {$u_1$}; 
			\node[user, yshift=-1.7cm, xshift=1cm] (2) {$u_2$}; 
			\node[user, yshift=-2.3cm, xshift=3.5cm] (3) {$u_4$}; 
			\node[user, xshift=2.5cm, yshift=-.5cm] (4) {$u_3$}; 
			\draw[<->] (1) to (2);
			\draw[<->] (2) to (3);
			\draw[<->] (2) to (4);
			\draw[<->] (4) to (3);
		\end{tikzpicture}
		}
    }
	\caption{Schematic representation of decentralized and federated learning. S stands for Server.}
	\label{fig:cml}
\end{figure}

%% file: images/so_tau.tex
\begin{figure*}[h]
    \centering
    \subfloat[Torus]{\label{fig:tau_torus}
        \includegraphics[width=0.33\linewidth]{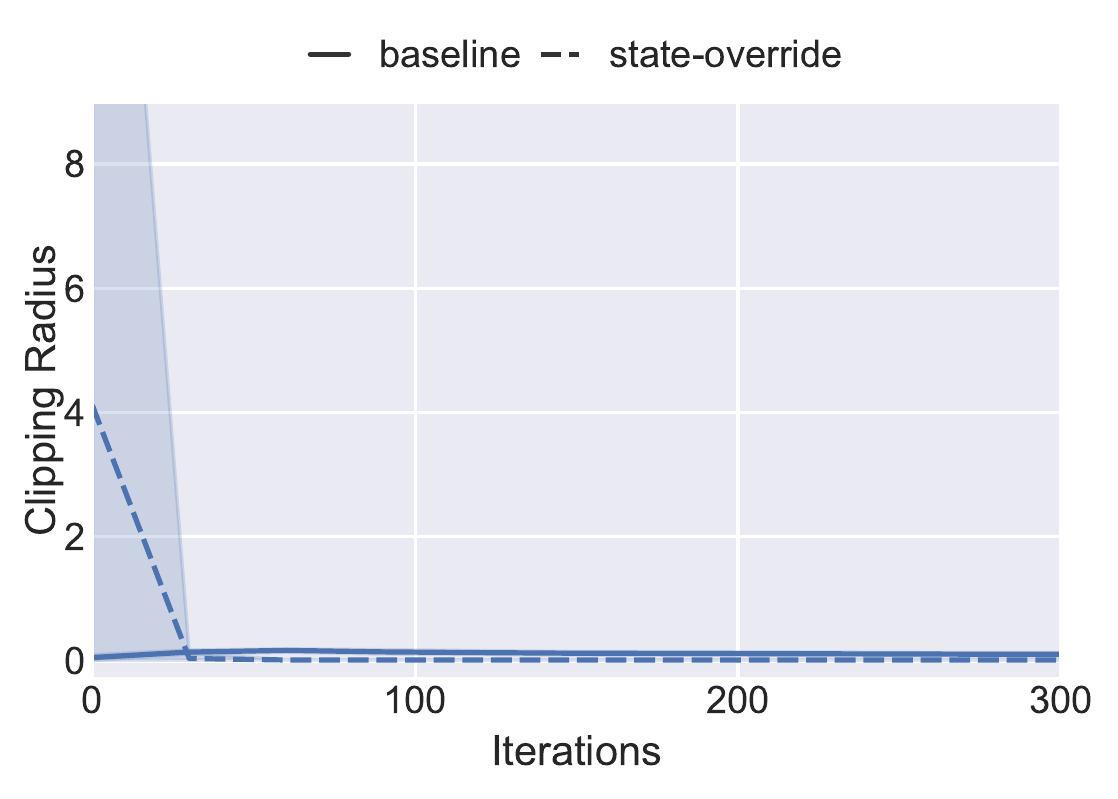}
    }\subfloat[Dumbbell]{\label{fig:tau_dumbbell}
        \includegraphics[width=0.33\linewidth]{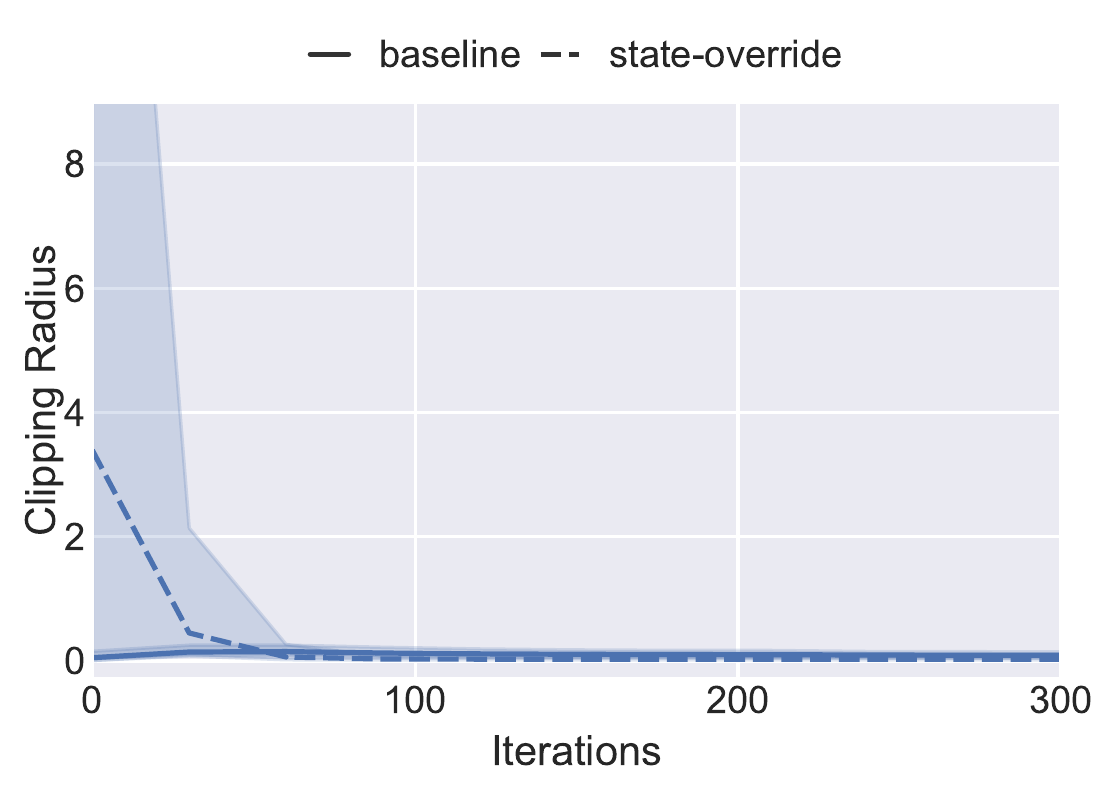}
    }
    \caption{Clipping radius in a benign training and under the state--override attack, using Eq. \ref{eq:tau}.}
    \label{fig:tau}
\end{figure*}

%% file: parts/5e-aggregators.tex
\section{Robust aggregators in FL}
\label{sec:ra}

In DL, the users receive only a subset of all the updates computed at the current iteration, and need to decide on their next local state using these updates only. 
When robust aggregation is used, this decision includes filtering updates that are not valuable for the learning task, e.g., adversarial.
However, any decision that appears to be suitable locally, i.e., based on the view of the user, might not be the best at a global scale.
In order to produce better decisions, equivalently, for the result of the aggregation to be more robust, it is beneficial to maximize the number of updates used to produce the aggregation result.
Eventually, when the user receives all updates, i.e., it is connected to all other users, it can produce the best decision. 

There might be better robust aggregators introduced in the future which improve on existing ones.
However, this will not give any advantage to DL.
Indeed, any robust aggregation technique from the DL literature~\cite{ogclipping, scc, jungle} can be implemented by the server in FL.
Since the server observes all updates, the robustness of the aggregator is maximized and matches the result of the aggregation performed by decentralized users in a complete topology.

\begin{figure}[h]
    \centering
    \includegraphics[width=0.43\linewidth]{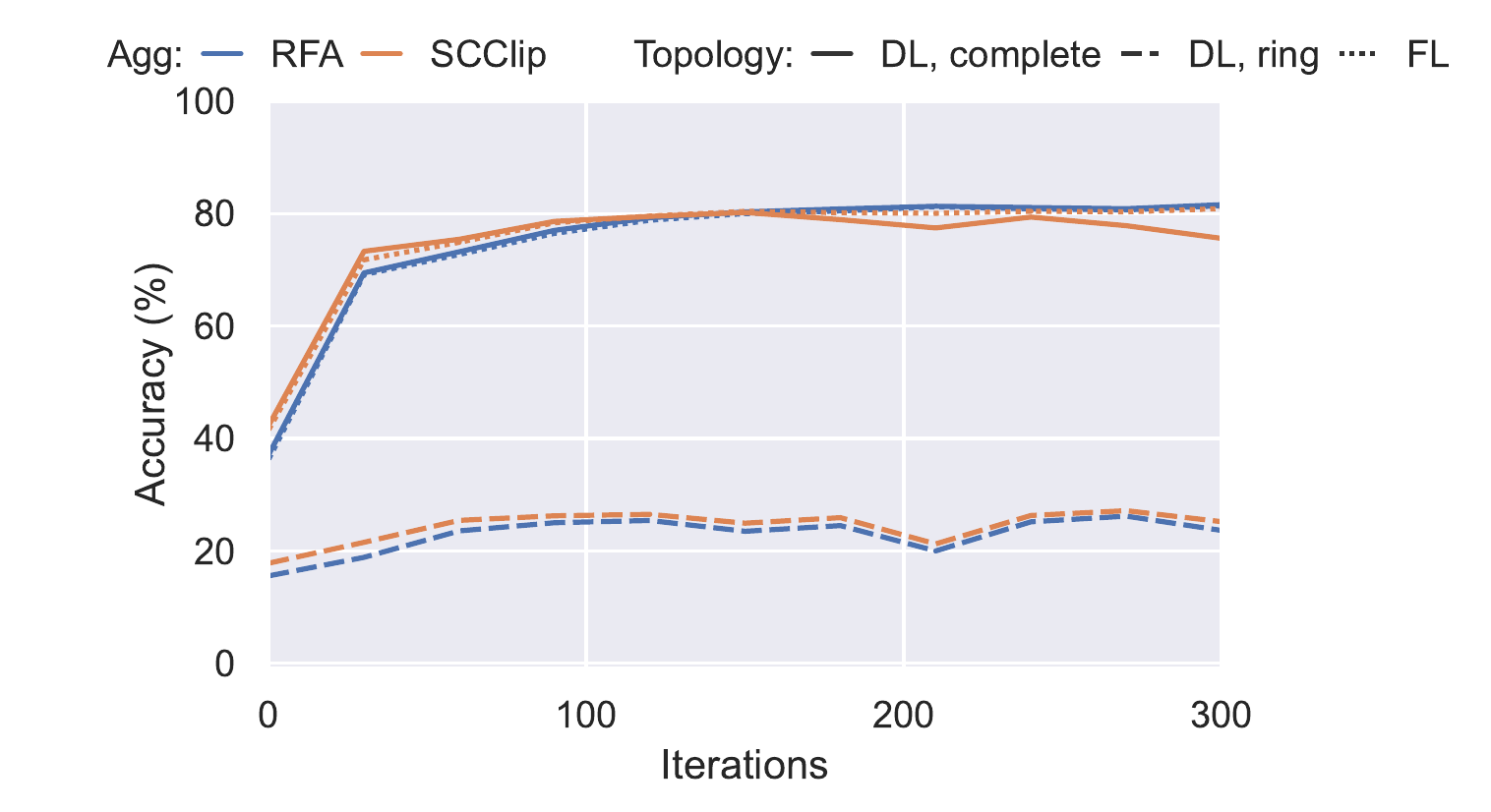}
    \caption{Accuracy of the model of the user $u_4$ under the noisy attack in DL and FL using robust aggregation.}
    \label{fig:ra}
\end{figure}

In Figure~\ref{fig:ra}, we report the robustness achieved in FL and DL, when the robust aggregator is respectively implemented at the server--side and the user--side.
We take the same experimental setup as in Section~\ref{sec:incons}: we consider 8 honest users and 1 Byzantine performing the noisy attack.
For this experiment, we use robust aggregation instead of naive.
We test two robust aggregators: RFA~\cite{rfa} and \SCClip{} with the ideal clipping radius.
We look at the robustness of the training in FL, and in DL in both the ring and the complete topologies.
Our first observation is that when robust aggregation is used, the noisy attack is ineffective in DL with complete topology (compared to Figure~\ref{fig:inconst}), i.e., the accuracy steadily remains above $75\%$.
We then notice that in the ring topology---where users only receive 2 updates---the user converge to a non--accurate model.
We conclude that both RFA and \SCClip{} provide the best robustness under the complete topology.
Finally, we observe that the accuracy, hence the robustness, achieved in DL with a complete topology and in FL is extremely close.
This demonstrates that these robust aggregators can be implemented by the server and provide robustness comparable to the one obtained in DL in the complete topology.

In conclusion, \textbf{robust aggregators are not sufficient to provide any advantage of DL over FL}.

